# Dynamic Magnetometer Calibration and Alignment to Inertial Sensors by Kalman Filtering

Yuanxin Wu, *Senior Member, IEEE*, Danping Zou, Peilin Liu and Wenxian Yu

*Abstract*— Magnetometer and inertial sensors are widely used for orientation estimation. Magnetometer usage is often troublesome, as it is prone to be interfered by onboard or ambient magnetic disturbance. The onboard soft-iron material distorts not only the magnetic field, but the magnetometer sensor frame coordinate and the cross-sensor misalignment relative to inertial sensors. It is desirable to conveniently put magnetic and inertial sensors information in a common frame. Existing methods either split the problem into successive intrinsic and cross-sensor calibrations, or rely on stationary accelerometer measurements which is infeasible in dynamic conditions. This paper formulates the magnetometer calibration and alignment to inertial sensors as a state estimation problem, and collectively solves the magnetometer intrinsic and cross-sensor calibrations, as well as the gyroscope bias estimation. Sufficient conditions are derived for the problem to be globally observable, even when no accelerometer information is used at all. An extended Kalman filter is designed to implement the state estimation and comprehensive test data results show the superior performance of the proposed approach. It is immune to acceleration disturbance and applicable potentially in any dynamic conditions.

## I. INTRODUCTION

Low-cost modules or chips, consisting of three triads of gyroscopes, accelerometers and magnetometers, become ubiquitous nowadays in consumer devices. They are used in unmanned aerial vehicles for flight stabilization, in mobile phones for personal navigation, in virtual reality helmets for attitude tracking and in wearable units for human body motion tracking. Gyroscopes, accelerometers and magnetometer are three different kinds of sensors with distinctive features, of which the former two are known as inertial sensors. Gyroscopes sense the body angular rate with respective to the inertial frame and the computed rotation by integration subjects to drift due to the presence of the gyroscope bias and noise. Accelerometers measure the non-gravitational acceleration that cannot be used to derive the inclination until they are roughly at rest. Magnetometers sense the local magnetic field that may be interfered by the nearby ferromagnetic material or strong electric currents. Specifically, the soft-iron material distorts not only the ambient magnetic field but the magnetometer coordinate frame. As a result, the cross-sensor frame misalignment between magnetometers and gyroscopes/accelerometers are altered as well, so it is advised that both intrinsic and cross-sensor calibrations be carried out prior to magnetometer usage [1]. In comparison, the coordinate frame of and the mutual frame between gyroscopes and accelerometers change little under normal working conditions and is usually calibrated once for all.

As the magnetometer frame is potentially affected by other sensors or the attached platform itself, the magnetometer calibration should not be performed until all sensors are rigidly fixed to the platform. The calibration process commonly consists of two steps: (1) intrinsic calibration and (2) cross-sensor calibration. The attitude-independent method [2-10] is the most popular magnetometer intrinsic calibration approach, which exploits the fact that the magnetometer measurement at the local position or in a homogeneous magnetic field is constant in magnitude regardless of the orientation. The cross-sensor misalignment of accelerometers and magnetometers is estimated in [11] using the invariant angle formed by the local gravity vector and the local magnetic field vector. The cross-sensor misalignment of magnetometers relative to gyroscopes is determined in [4] by using the gyroscope-derived incremental rotation as a reference, and in [1] by exploiting the fact that in a homogenous magnetic field the magnetometer's measurement variation is exclusively induced by orientation change. The incremental rotation of the magnetometer is assumed known in [12], which restricts its usage in controlled environments with attitude reference only. In [7, 13, 14], the intrinsic calibration and cross-sensor calibration are collectively addressed as a maximum likelihood estimation using magnetometer and accelerometer measurements. Almost all previous works, e.g. [4, 7, 11-14], rely on the local gravity information for the cross-sensor calibration, thus requiring to collect accelerometer measurements at stationary poses, and any disturbed acceleration would inevitably decay the calibration quality. An exception is our work [1] that solves the cross-sensor misalignment by the recursive optimization using only magnetometer and gyroscope measurements, which is immune to any acceleration disturbance. As the intrinsic calibration has to be carried out a prior using the whole dataset, however, the recursive attribute of the proposed cross-sensor calibration solution [1] is significantly compromised.

Compared with all previous works, the main contribution of this paper is a novel state estimation approach collectively solving the magnetometer intrinsic and cross-sensor calibrations in a homogeneous magnetic field. With respect to our work [1], the proposed approach is truly recursive in time and immune to acceleration disturbance. The paper is organized as follows. Section II describes the sensor models of gyroscope, accelerometer and magnetometer. Magnetometer

This work was supported in part by National Natural Science Foundation of China (61422311, 61673263) and Hunan Provincial Natural Science Foundation of China (2015JJ1021).

Authors' address: Shanghai Key Laboratory of Navigation and Location-based Services, School of Electronic Information and Electrical Engineering, Shanghai Jiao Tong University, Shanghai, China, 200240, E-mail: (yuanx_wu@hotmail.com).

calibration/alignment is formulated as a state estimation problem and two sufficient conditions of global observability are derived. Section III uses extended Kalman filtering to solve the problem and reports the field test results. The conclusions are drawn in Section V.

## II. SENSOR MODELS AND MAGNETOMETER CALIBRATION PROBLEM

### A. Measurement Models: Magnetometer, Accelerometer and Gyroscope

Assuming the magnetic disturbance is time-invariant and taking sensor imperfection into account, the magnetometer measurement can be modelled by [2, 4]

$$\mathbf{y}_m = \mathbf{S}\mathbf{C}_e^b \mathbf{m}^e + \mathbf{h} + \mathbf{n}_m \quad (1)$$

where $\mathbf{m}^e$ is the local magnetic vector in the Earth frame (e-frame) and $\mathbf{n}_m$ is i.i.d zero-mean Gaussian noise with covariance $\sigma_m^2 \mathbf{I}_3$. In a homogeneous magnetic field, $\mathbf{m}^e$ is constant and assumed to have unity norm without loss of generality. The body frame (b-frame) refers to the coordinate frame defined by the gyroscope/accelerometer triads. The attitude or orientation matrix $\mathbf{C}_e^b$ transforms the magnetic field vector from e-frame to b-frame. The purpose of the magnetometer calibration is to determine the matrix $\mathbf{S}$ and the vector $\mathbf{h}$, which collectively encode the magnetometer sensor triad imperfection, magnetic disturbance, and the misalignment with respect to b-frame. The calibration matrix $\mathbf{S}$ can be decomposed as $\mathbf{S}^{-1} = \mathbf{C}_m^b \mathbf{R}$ by the orthogonal-triangular decomposition, where $\mathbf{C}_m^b$ is the cross-sensor misalignment between the magnetometer frame (m-frame) and the body frame and $\mathbf{R}$ (with positive diagonal entries) belongs to the magnetometer intrinsic parameters. As shown in [1], m-frame is typically distinctive from the physical coordinate frame of the magnetic sensor triad in the presence of soft-iron materials. Equation (1) is a general magnetometer calibration model that combines together the intrinsic calibration and the cross-sensor calibration. For example, the attitude-independent intrinsic calibration method [2-10] aims to seek the parameters $\mathbf{R}$ and $\mathbf{h}$ that satisfy $\|\mathbf{R}(\mathbf{y}_m - \mathbf{h} - \mathbf{n}_m)\| = \|\mathbf{C}_b^m \mathbf{C}_e^b \mathbf{m}^e\| = 1$.

The accelerometer measurement is related to the local gravity vector by [15, 16]

$$\mathbf{y}_a = \mathbf{C}_e^b \left(-\mathbf{g}^e + \dot{\mathbf{v}}^e + 2\boldsymbol{\omega}_{ie}^e \times \mathbf{v}^e\right) + \mathbf{n}_a \approx -\mathbf{C}_e^b \mathbf{g}^e + \mathbf{n}_a \quad (2)$$

where $\mathbf{g}^e$ is the true local constant gravity vector in e-frame, $\dot{\mathbf{v}}^e + 2\boldsymbol{\omega}_{ie}^e \times \mathbf{v}^e$ is the sum of linear and Coriolis accelerations and $\mathbf{n}_{aw}$ is i.i.d Gaussian noise. $\mathbf{v}^e$ is the velocity relative to the Earth and $\boldsymbol{\omega}_{ie}^e$ is the Earth's rotation rate. When the acceleration sum term was restrained to be small, the accelerometer measurement $\mathbf{y}_a$ would roughly reflect the local gravity and for this situation $\mathbf{n}_a$ could be modelled as zero-mean noise with covariance $\sigma_a^2 \mathbf{I}_3$. As shown in (2), the accelerometer is usually considered as ideal for low-cost applications as its error has much less impact on orientation than the gyroscope.

Equations (1)-(2) indicate that the magnetometer and accelerometer measurements are both related to the body attitude matrix $\mathbf{C}_e^b$ which projects the local magnetic vector and the local gravity vector to b-frame. By the chain rule of the orientation matri

$$\mathbf{C}_e^b(t) = \mathbf{C}_{b(0)}^{b(t)} \mathbf{C}_e^b(0) \mathbf{C}_{e(t)}^{e(0)} \quad (3)$$

in which $\mathbf{C}_{b(0)}^{b(t)}$ and $\mathbf{C}_{e(t)}^{e(0)}$ respectively describe the orientation change of b-frame and e-frame over the considered time period. Their time rates are related to the corresponding angular velocities by [15]

$$\begin{aligned}\dot{\mathbf{C}}_{b(t)}^{b(0)} &= \mathbf{C}_{b(t)}^{b(0)} \left(\boldsymbol{\omega}_{ib}^b - \boldsymbol{\varepsilon} - \mathbf{n}_g\right)\times \\ \dot{\mathbf{C}}_{e(t)}^{e(0)} &= \mathbf{C}_{e(t)}^{e(0)} \left(\boldsymbol{\omega}_{ie}^e \times\right)\end{aligned} \quad (4)$$

where $\boldsymbol{\omega}_{ib}^b$ denotes the gyroscope measurement, $\boldsymbol{\varepsilon}$ the gyroscope bias and $\mathbf{n}_g$ i.i.d zero-mean Gaussian noise with covariance $\sigma_g^2 \mathbf{I}_3$. The skew symmetric matrix $(\cdot \times)$ is defined so that the cross product satisfies $\mathbf{x} \times \mathbf{y} = (\mathbf{x} \times) \mathbf{y}$ for arbitrary two vectors. For any constant vector $\mathbf{r}^e$,

$$\mathbf{C}_e^b \mathbf{r}^e = \mathbf{C}_{b(0)}^{b(t)} \mathbf{C}_e^b(0) \mathbf{C}_{e(t)}^{e(0)} \mathbf{r}^e = \mathbf{C}_{b(0)}^{b(t)} \mathbf{r}^{b(0)} \quad (5)$$

Using (4), the magnitude of the vector rate $\|\dot{\mathbf{r}}^{b(0)}\| = \|\mathbf{C}_e^b(0) \mathbf{C}_{e(t)}^{e(0)} (\boldsymbol{\omega}_{ie}^e \times \mathbf{r}^e)\| < \Omega \|\mathbf{r}^e\|$. As the magnitude of the Earth's rotation rate ($\Omega \approx 7.3 \times 10^{-5}$ rad/s) is much smaller than low-cost gyroscope errors, it is reasonable to take $\mathbf{r}^{b(0)}$ approximately as a constant.

### B. Magnetometer Calibration and Observability Property

Hereafter we use the initial body frame, namely $b(0)$, as the inertial frame (i-frame). The above equations can be formulated in a state-space form as

$$\begin{aligned}\dot{\mathbf{C}}_{b(t)}^i &= \mathbf{C}_{b(t)}^i \left(\boldsymbol{\omega}_{ib}^b - \boldsymbol{\varepsilon} - \mathbf{n}_g\right)\times \\ \dot{\boldsymbol{\varepsilon}} &= \mathbf{n}_\varepsilon \\ \dot{\mathbf{S}} &= 0, \quad \dot{\mathbf{h}} = 0 \\ \dot{\mathbf{m}}^i &= \mathbf{n}_{mi}, \quad \dot{\mathbf{g}}^i = \mathbf{n}_{gi}\end{aligned} \quad (6)$$

with the magnetometer and accelerometer measurements given as

$$\begin{aligned}\mathbf{y}_m &= \mathbf{S}\mathbf{C}_i^{b(t)} \mathbf{m}^i + \mathbf{h} + \mathbf{n}_m \\ \mathbf{y}_a &= -\mathbf{C}_i^{b(t)} \mathbf{g}^i + \mathbf{n}_a\end{aligned} \quad (7)$$

where $\mathbf{n}_\varepsilon$, $\mathbf{n}_{mi}$ and $\mathbf{n}_{gi}$ are i.i.d zero-mean Gaussian noises with covariance $\sigma_\varepsilon^2 \mathbf{I}_3$, $\sigma_{mi}^2 \mathbf{I}_3$ and $\sigma_{gi}^2 \mathbf{I}_3$, respectively. The augmented system state $\mathbf{x}$ comprises the inertial attitude $\mathbf{C}_i^{b(t)}$, the gyroscope bias $\boldsymbol{\varepsilon}$, the magnetometer calibration parameters $\mathbf{S}$ and $\mathbf{h}$, and the approximately constant vectors $\mathbf{m}^i$ and $\mathbf{g}^i$.

*Definition of State Observability* [17]: A system is said to be (globally) observable if for any unknown initial state $\mathbf{x}(0)$, there exists a finite $t > 0$ such that the knowledge of the input and the output over $[0, t]$ suffices to determine uniquely the initial state $\mathbf{x}(0)$. Otherwise, the system is said to be (globally) unobservable.

This is a concept of deterministic observability taking no account of noises. Whatever estimation techniques are to be used, observability analysis is necessary that tells the inherent estimability of the system state [17, 18].

*Theorem 2.1*: If the matrix $\int_0^t \mathbf{W}^T \mathbf{W} dt$ is nonsingular, the matrix $\int_0^t \mathbf{Y}^{*T} \mathbf{Y}^* dt$ has one and only one zero eigenvalue and $\mathbf{C}_i^{b(0)}$ is restricted to be an identical matrix, then the system state is globally observable. ($\mathbf{Y}^*$ and $\mathbf{W}$ are defined as in (9) and (12) below)

Proof. Making use of the unity-norm property of $\mathbf{m}^i$ and the orthogonal-triangular decomposition $\mathbf{S}^{-1} = \mathbf{C}_m^b \mathbf{R}$, the magnetometer measurement in (7) yields

$$1 = \|\mathbf{m}^i\| = \|\mathbf{C}_{b(t)}^i \mathbf{C}_m^b \mathbf{R}(\mathbf{y}_m - \mathbf{h})\| = \|\mathbf{R}(\mathbf{y}_m - \mathbf{h})\| \quad (8)$$

Expanding the above equation and expressing it in a linear equation form [7, 8]

$$\begin{bmatrix} \mathbf{y}_m^T \otimes \mathbf{y}_m^T & -2\mathbf{y}_m^T & 1 \end{bmatrix} \begin{bmatrix} vec(\mathbf{R}^T \mathbf{R}) \\ \mathbf{R}^T \mathbf{R} \mathbf{h} \\ \mathbf{h}^T \mathbf{R}^T \mathbf{R} \mathbf{h} - 1 \end{bmatrix} \equiv \mathbf{Y} \mathbf{z} = 0 \quad (9)$$

The operator $\otimes$ denotes the Kronecker product. As $\mathbf{R}^T \mathbf{R}$ is symmetric, $vec(\mathbf{R}^T \mathbf{R})$ is formed by stacking the columns of $\mathbf{R}^T \mathbf{R}$ but excluding the lower triangular entries. The columns of $\mathbf{Y}$ corresponding to the three lower triangular entries are merged to those columns corresponding to their symmetric counterparts, so are the columns of $\mathbf{z}$. To avoid notation confusion, we denote the modification by superscript asterisk, namely, (9) becomes $\mathbf{Y}^* \mathbf{z}^* = 0$. Left multiplying $\mathbf{Y}^{*T}$ and integrating over the interested time interval, it gives $\int_0^t \mathbf{Y}^{*T} \mathbf{Y}^* dt \, \mathbf{z}^* = 0$. If the matrix $\int_0^t \mathbf{Y}^{*T} \mathbf{Y}^* dt$ has only one zero eigenvalue, then the solution of $\mathbf{z}^*$ is the corresponding eigenvector, from which $\mathbf{R}$ and $\mathbf{h}$ can be uniquely determined.

Denote $\mathbf{y}_m^* = \mathbf{R}(\mathbf{y}_m - \mathbf{h})$. The magnetometer measurement in (7) becomes

$$\mathbf{y}_m^* = \mathbf{C}_b^m \mathbf{C}_i^{b(t)} \mathbf{m}^i \quad (10)$$

Taking time derivative and using (4),

$$\dot{\mathbf{y}}_m^* = -\mathbf{C}_b^m \left( (\boldsymbol{\omega}_{ib}^b - \boldsymbol{\varepsilon}) \times \right) \mathbf{C}_i^{b(t)} \mathbf{m}^i = (\mathbf{y}_m^* \times) \mathbf{C}_b^m (\boldsymbol{\omega}_{ib}^b - \boldsymbol{\varepsilon}) \quad (11)$$

or equivalently,

$$\dot{\mathbf{y}}_m^* = \begin{bmatrix} \boldsymbol{\omega}_{ib}^{bT} \otimes (\mathbf{y}_m^* \times) & -(\mathbf{y}_m^* \times) \end{bmatrix} \begin{bmatrix} vec(\mathbf{C}_b^m) \\ \mathbf{C}_b^m \boldsymbol{\varepsilon} \end{bmatrix} \equiv \mathbf{W} \boldsymbol{\eta} \quad (12)$$

If the matrix $\int_0^t \mathbf{W}^T \mathbf{W} dt$ is nonsingular, then $\boldsymbol{\eta}$ can be solved as $\boldsymbol{\eta} = \left( \int_0^t \mathbf{W}^T \mathbf{W} dt \right)^{-1} \int_0^t \mathbf{W}^T \dot{\mathbf{y}}_m^* dt$, from which the misalignment $\mathbf{C}_b^m$ and the gyroscope bias $\boldsymbol{\varepsilon}$ will be determined. Note that this nonsingular condition is sufficient but not necessary, as it does not consider the constraint among the entities of an orientation matrix.

However, the initial values for $\mathbf{C}_i^{b(t)}$ and $\mathbf{m}^i$ cannot be uniquely determined, as for any attitude matrix $\mathbf{Q}$ the following equation is always valid for the magnetometer measurement in (7)

$$\mathbf{y}_m = \mathbf{S}(\mathbf{C}_i^{b(t)} \mathbf{Q})(\mathbf{Q}^T \mathbf{m}^i) + \mathbf{h} \quad (13)$$

which means that the initial values of $\mathbf{C}_i^{b(t)}$ and $\mathbf{m}^i$ both have infinite feasible solutions. Therefore, we restrict $\mathbf{C}_i^{b(t)}\big|_{t=0}$ to be its physical value, namely an identical matrix, to make the formulation (6)-(7) fully observable. Then the inertial attitude $\mathbf{C}_i^{b(t)}$ will be available with (4) and the determined gyroscope bias through integration. Then the constant magnetic vector is computed as $\mathbf{m}^i = \mathbf{C}_{b(t)}^i \mathbf{S}^{-1}(\mathbf{y}_m - \mathbf{h})$ and the constant gravity vector is computed as $\mathbf{g}^i = -\mathbf{C}_{b(t)}^i \mathbf{y}_a$ from the accelerometer measurement equation in (7). ∎

The observability analysis in Theorem 2.1 overwhelmingly depends on the magnetometer/gyroscope measurement information, and the subsequent result provides another set of sufficient conditions for the system to be observable by additionally exploiting the accelerometer information.

*Theorem 2.2*: If the matrix $\int_0^t \mathbf{y}_a \mathbf{y}_a^T dt$ is nonsingular, the matrix $\int_0^t \mathbf{M}^T \mathbf{M} dt$ has one and only one zero eigenvalue and $\mathbf{C}_i^{b(0)}$ is restricted to be an identical matrix, then the system state is globally observable. ($\mathbf{M}$ is defined as in (17) below)

Proof. Taking the time derivative of the accelerometer measurement in (7) and using (4),

$$\dot{\mathbf{y}}_a = \left( (\boldsymbol{\omega}_{ib}^b - \boldsymbol{\varepsilon}) \times \right) \mathbf{C}_i^{b(t)} \mathbf{g}^i = -(\boldsymbol{\omega}_{ib}^b - \boldsymbol{\varepsilon}) \times \mathbf{y}_a \quad (14)$$

from which the gyroscope bias, if the matrix $\int_0^t \mathbf{y}_a \mathbf{y}_a^T dt$ is nonsingular, can be uniquely solved as $\boldsymbol{\varepsilon} = -\left( \int_0^t \mathbf{y}_a \mathbf{y}_a^T dt \right)^{-1} \int_0^t (\dot{\mathbf{y}}_a + \boldsymbol{\omega}_{ib}^b \times \mathbf{y}_a) dt$. Similarly the initial values for $\mathbf{C}_i^{b(t)}$ and $\mathbf{g}^i$ have infinite feasible solutions, because for any attitude matrix $\mathbf{Q}$ the following equation is always valid

$$\mathbf{y}_a = -(\mathbf{C}_i^{b(t)} \mathbf{Q})(\mathbf{Q}^T \mathbf{g}^i) \quad (15)$$

By restricting $\mathbf{C}_i^{b(0)}$ to be an identical matrix, the inertial attitude $\mathbf{C}_i^{b(t)}$ will be available with (4) and the determined gyroscope bias. With the known inertial attitude $\mathbf{C}_i^{b(t)}$, the constant gravity vector is computed as $\mathbf{g}^i = -\mathbf{C}_{b(t)}^i \mathbf{y}_a$ and the magnetometer measurement equation in (7) can be re-organized as

$$\mathbf{C}_{b(t)}^i \mathbf{S}^{-1} \mathbf{y}_m - \mathbf{C}_{b(t)}^i \mathbf{S}^{-1} \mathbf{h} - \mathbf{m}^i = 0 \quad (16)$$

or alternatively in the linear equation form

$$\begin{bmatrix} \mathbf{y}_m^T \otimes \mathbf{C}_{b(t)}^i & -\mathbf{C}_{b(t)}^i & -\mathbf{I}_3 \end{bmatrix} \begin{bmatrix} vec(\mathbf{S}^{-1}) \\ \mathbf{S}^{-1}\mathbf{h} \\ \mathbf{m}^i \end{bmatrix} \equiv \mathbf{M}\boldsymbol{\kappa} = 0 \quad (17)$$

If the matrix $\int_0^t \mathbf{M}^T\mathbf{M}\,dt$ has only one zero eigenvalue, then the solution of $\boldsymbol{\kappa}$ is the corresponding eigenvector, from which the magnetometer parameters $\mathbf{S}$ and $\mathbf{h}$ and the constant magnetic vector $\mathbf{m}^i$ can be uniquely determined.

∎

## III. ESTIMATION ALGORITHM AND TEST RESULTS

### A. Estimation Algorithm

The discrete-time form of (6) is straightforward, as all sub-states but the inertial attitude are just copied from the last epoch $t_{k-1}$ to the current epoch $t_k$ and the inertial attitude is propagated to the first order as such [15, 16]

$$\begin{aligned} \mathbf{C}_{b(t_k)}^i &= \mathbf{C}_{b(t_{k-1})}^i \mathbf{C}_{b(t_k)}^{b(t_{k-1})} \\ &= \mathbf{C}_{b(t_{k-1})}^i \left[ \mathbf{I}_3 + (t_k - t_{k-1})(\boldsymbol{\omega}_{ib}^b - \boldsymbol{\varepsilon}_{k-1}) \times \right] \end{aligned} \quad (18)$$

where $\boldsymbol{\varepsilon}_{k-1}$ is the gyroscope bias estimate at the last epoch. The inertial attitude starts from an identical matrix with zero covariance, namely implementing the requirement of $\mathbf{C}_i^{b(0)}$ to be an identical matrix in Theorems 2.1-2.2. The matrix parameter $\mathbf{S}$ starts from an identical matrix as well, and the initial gyroscope bias $\boldsymbol{\varepsilon}$ and the initial magnetometer parameter $\mathbf{h}$ are both set to zeros. The initial values of $\mathbf{m}^i$ and $\mathbf{g}^i$ are respectively set to the magnetometer's and accelerometer's first measurements. Note that because of the potentially significant soft-iron effect, the above initial value setting might not be satisfactory for $\mathbf{S}$, $\mathbf{h}$ and $\mathbf{m}^i$. Some ad-hoc techniques might be needed, as discussed in next subsection.

The error-state extended Kalman filter (EKF) is employed to carry out the state estimation [19]. Deriving the first order error-state equation of the formulation (6)-(7) is straightforward and we directly present it below. Readers may refer to [15, 16] for details. Define the error state as the estimate subtracting the true state, i.e., $\delta\mathbf{x} = \hat{\mathbf{x}} - \mathbf{x}$. The attitude estimate is defined as being related to the true attitude and the corresponding attitude error $\boldsymbol{\psi}$ by $\hat{\mathbf{C}}_{b(t)}^i \approx (\mathbf{I} - \boldsymbol{\psi} \times)\mathbf{C}_{b(t)}^i$. The error state $\delta\mathbf{x} \equiv \begin{bmatrix} \boldsymbol{\psi}^T & \delta\boldsymbol{\varepsilon}^T & vec^T(\delta\mathbf{S}) & \delta\mathbf{h}^T & \delta\mathbf{m}^{iT} & \delta\mathbf{g}^{iT} \end{bmatrix}^T$, whose dynamic equation in the state-space matrix form is given by

$$\begin{cases} \delta\dot{\mathbf{x}} = \mathbf{F}\delta\mathbf{x} + \mathbf{G}\mathbf{w} \\ \delta\mathbf{y} \equiv \begin{bmatrix} \delta\mathbf{y}_m \\ \delta\mathbf{y}_a \end{bmatrix} = \begin{bmatrix} \mathbf{H}_m \\ \mathbf{H}_a \end{bmatrix} \delta\mathbf{x} + \begin{bmatrix} \mathbf{n}_m \\ \mathbf{n}_a \end{bmatrix} \end{cases} \quad (19)$$

where the dynamic noise $\mathbf{w} \equiv \begin{bmatrix} \mathbf{n}_g^T & \mathbf{n}_\varepsilon^T & \mathbf{n}_{mi}^T & \mathbf{n}_{gi}^T \end{bmatrix}^T$ and the matrices are

$$\mathbf{F} = \begin{bmatrix} \mathbf{0}_{3\times3} & -\mathbf{C}_b^i & \mathbf{0}_{3\times18} \\ \mathbf{0}_{21\times3} & \mathbf{0}_{21\times3} & \mathbf{0}_{21\times18} \end{bmatrix}, \quad \mathbf{G} = \begin{bmatrix} -\mathbf{C}_b^i & \mathbf{0}_{3\times3} & \mathbf{0}_{3\times6} \\ \mathbf{0}_{3\times3} & \mathbf{I}_3 & \mathbf{0}_{3\times6} \\ \mathbf{0}_{12\times3} & \mathbf{0}_{12\times3} & \mathbf{0}_{12\times6} \\ \mathbf{0}_{6\times3} & \mathbf{0}_{6\times3} & \mathbf{I}_6 \end{bmatrix},$$

$$\mathbf{H}_m = \begin{bmatrix} -\mathbf{S}\mathbf{C}_i^b(\mathbf{m}^i)\times & \mathbf{0}_{3\times3} & (\mathbf{C}_i^b\mathbf{m}^i)^T \otimes \mathbf{I}_3 & \mathbf{I}_3 & \mathbf{S}\mathbf{C}_i^b & \mathbf{0}_{3\times3} \end{bmatrix},$$

$$\mathbf{H}_a = \begin{bmatrix} \mathbf{C}_i^b\mathbf{g}^i \times & \mathbf{0}_{3\times3} & \mathbf{0}_{3\times9} & \mathbf{0}_{3\times3} & \mathbf{0}_{3\times3} & -\mathbf{C}_i^b \end{bmatrix}$$
(20)

In a homogeneous magnetic field, the magnetometer measurement can always be used to update the state estimate, but the accelerometer measurement's usefulness depends. We adopt a simple thresholding scheme that only those accelerometer measurements approximately equal to the gravity vector in magnitude are accepted for the EKF update. Specifically, the valid accelerometer measurements should satisfy $\|\mathbf{y}_a\| - g| < T_{md}$, where $T_{md}$ is a prescribed threshold of magnitude discrepancy and $g$ is the local gravity magnitude [20, 21].

Note that the state of the constant magnetic vector $\mathbf{m}^i$ should have unity magnitude. We have tried the equality-constrained EKF techniques proposed in [22, 23] but they performed unsatisfactory in this problem. The following technique is found to quite effective. In fact, if the unit-magnitude constraint of the constant magnetic vector was not considered in the EKF, (7) indicates that there would only exist a scalar ambiguity between the magnetometer matrix $\mathbf{S}$ and the magnetic vector $\mathbf{m}^i$, namely $\alpha\mathbf{S}$ and $\alpha^{-1}\mathbf{m}^i$ for any nonzero scalar $\alpha$. Therefore, in this paper we disregard the norm constraint in the EKF, and after the EKF run re-scale the obtained estimates of the magnetometer matrix and the magnetic vector by $\mathbf{S}_{rs} = \|\mathbf{m}^i\|\mathbf{S}$ and $\mathbf{m}_{rs}^i = \mathbf{m}^i/\|\mathbf{m}^i\|$. Note that this has no effect on the observability analysis, as the

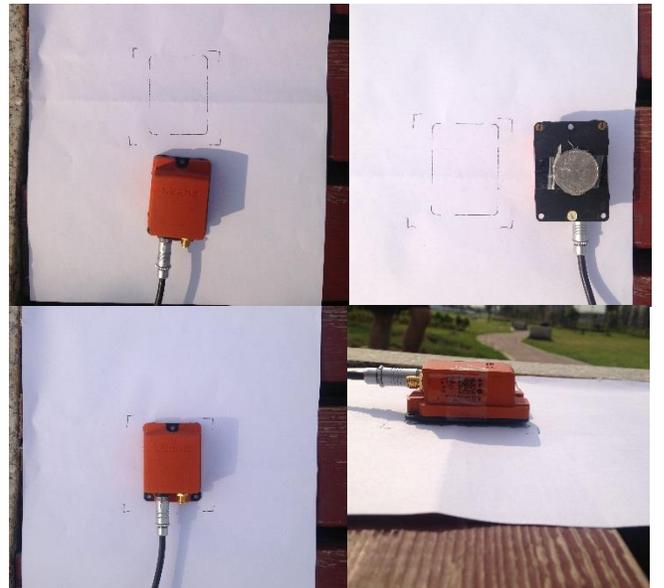

Figure 1. Test unit (Xsens MTi-G-700) and placement as indicated by rectangular (upper-right: a coin attached at unit bottom in Tests #3-4)

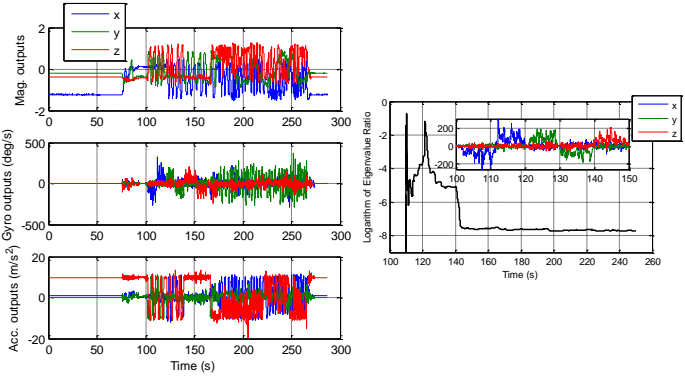

Figure 2. Magnetometer, gyroscope and accelerometer outputs in Test #1. (motion sequence: still at placement, picked up, short still, tumbled, short still, put back, still at same placement)

Figure 3. Eigenvalue ratio (smallest over second smallest) logarithm of $\int_0^t \mathbf{Y}^{*T}\mathbf{Y}^* dt$ for Test #1

nonzero scalar can always be cancelled from both sides of (8).

### B. Test Results

This subsection is devoted to verifying the above analysis and algorithm by real tests. We used the four test datasets in [1] that were collected using the Xsens MTi-G-700 unit. Each test starts and ends at an exactly same location on a wood bench, as shown by the rectangular in Fig. 1. Two datasets (#1 and #2) were collected using the MTi-G-700 unit alone, while the other two datasets (#3 and #4) were collected with a RMB coin taped onto the unit bottom plate. The coin is made of soft-iron magnetic material (Fig. 1, upper-right). The unit outputs are sampled at 100Hz. The tests were performed at latitude 28.247 deg and longitude 113.017 deg, in June 2015. The algorithm was run on a personal computer.

The magnetometer, gyroscope and accelerometer outputs in Test #1 are plotted in Fig. 2 as an example. The unit stays stationary on the bench for about 75 seconds, and is picked up and taken several meters away from the data-collecting desktop. After being held still for a moment, the unit starts to be tumbled by hand for about 170 seconds. Then the test is finalized by holding the unit still for a moment again and putting it back to the placement on the bench. The other datasets follow the same motion sequence. The estimation starts at 110s instead of at some earlier time, because the unit is stationary at about 90s-100s when the problem is obviously unobservable. Figure 3 gives the ratio of the smallest eigenvalue over the second smallest eigenvalue for the matrix $\int_0^t \mathbf{Y}^{*T}\mathbf{Y}^* dt$ in the logarithm scale. The eigenvalue ratio reaching almost zero clearly indicates the uniqueness of zero eigenvalue. Note that the ratio has two sharp reductions at 120s and 140s which respectively correspond to the starting time of the rotations along the y-axis and z-axis in the zoomed-in window of gyroscope measurements in Figure 2. This observation is common for all datasets.

Table I lists the gyroscope biases obtained by averaging the stationary data before motion in all tests. The relative attitude in Test #1 computed by integrating the gyroscope measurements after subtracting the calculated bias is plotted in Figure 4. As the unit is put at the same placement before and after the test, the end angles should ideally be zeros and the offsets (less than 2 degrees in about five minutes) reflect the

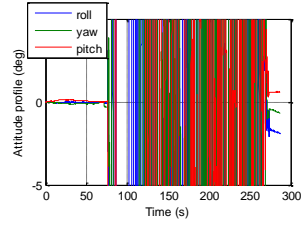
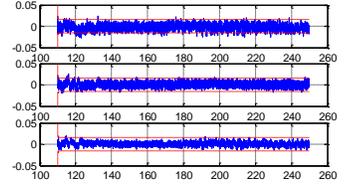

Figure 4. Attitude profile in Test #1, computed by integrating gyroscope measurements (still-averaging bias removed)

Figure 5. Magnetometer measurement innovation and three times EKF-computed standard variance (dashed line) for Test #1

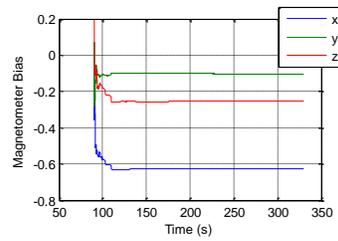
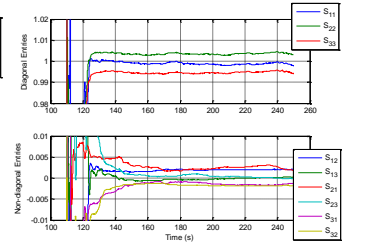

Figure 6. Gyroscope bias estimate for Test #1

Figure 7. Scaled magnetometer matrix estimate for Test #1

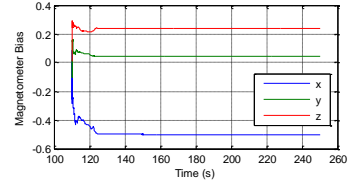
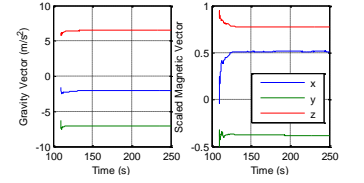

Figure 8. Magnetometer bias estimate for Test #1

Figure 9. Gravity vector estimate and scaled magnetic vector estimate for Test #1. Gravity vector has three nonzero coordinates due to special selection of inertial frame

accumulating errors mainly due to gyroscope noise and the neglected Earth rotation.

The noise standard variances in (19) are set by specifications or by examining the stationary data statistics as $\sigma_g = 0.01 \deg/\sqrt{s}$, $\sigma_\varepsilon = 10^{-4} \deg/\sqrt{s^3}$, $\sigma_{mi} = \Omega$, $\sigma_{gi} = 9.8\Omega\, m/\sqrt{s^5}$, $\sigma_m = 0.005$ and $\sigma_a = 3T_{md}$. The magnetic noise standard variances $\sigma_{mi}$ and $\sigma_m$ are unit-less as the true magnetometer measurement is assumed having unit magnitude, and $\sigma_{mi}$ and $\sigma_{gi}$ are set according to the analysis below (5). The magnitude discrepancy threshold $T_{md} = 0.03\, m/s^2$, and the accelerometer measurement noise standard variance is set to three times the threshold $T_{md}$ to account for the un-modeled acceleration disturbance. The initial state standard variance setting uses 5 deg/s for the gyroscope bias, 0.1 for the magnetometer matrix, 1 for the magnetometer bias, 0.5 for the magnetic vector and 1 m/s² for the gravity vector.

The EKF algorithm is applied to the hand-tumbling data in Test #1 (110s-250s). Figure 5 gives the magnetometer measurement innovation and the EKF-computed three times standard variance. The average normalized innovation squared (ANIS) across all time [24], namely the average of $(\mathbf{y} - \mathbf{y}_{k|k-1})^T (\mathbf{H}\mathbf{P}_{k|k-1}\mathbf{H}^T + \mathbf{R})^{-1} (\mathbf{y} - \mathbf{y}_{k|k-1})$, is employed as a scalar metric to quantify the measurement innovation, where $y_{k|k-1}$ is the predicted measurement, $P_{k|k-1}$ is the predicted state covariance and $H$ is the measurement matrix. Statistically, ANIS is chi-square distributed with the same degrees of freedom as the measurement $y$. For Test #1, the magnetometer ANIS is calculated to be 3.56, close to the theoretical value 3.

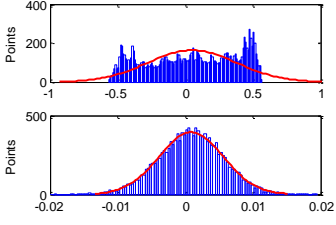
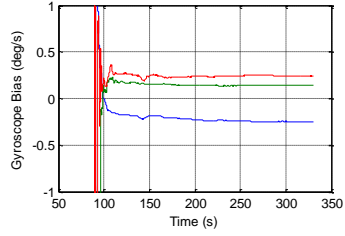

Figure 10. Histogram of magnitude discrepancy and fitted normal contribution (before and after applying magnetometer parameters by EKF for hand-tumbling data in Test #1)

Figure 11. Gyroscope bias estimate for Test #3 (2nd EKF run)

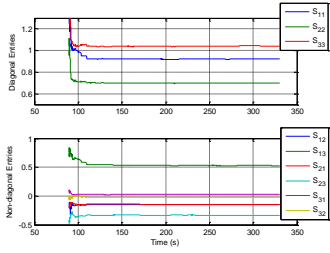
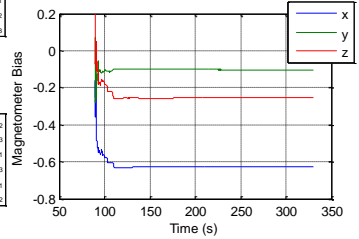

Figure 12. Scaled magnetometer matrix estimate for Test #3 (2nd EKF run)

Figure 13. Magnetometer bias estimate for Test #3 (2nd EKF run)

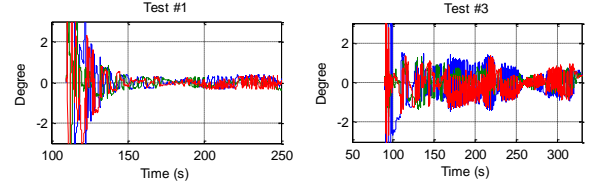

Figure 14. Euler angle discrepancy between inertial attitude estimate and gyroscope-maintained relative attitude for Test #1 and Test #3 (2nd EKF run)

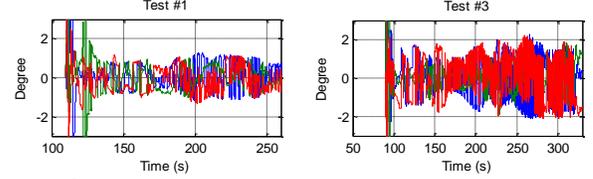

Figure 15. Euler angle discrepancy between inertial attitude estimate and gyroscope-maintained relative attitude for Test #1 and Test #3 (2nd EKF run) when accelerometer not used

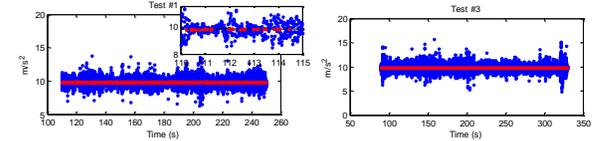

Figure 16. Magnitude of accelerometer outputs and those accepted for EKF update (in red dots) in Tests #1 and #3. In left-upper zoomed-in subfigure, black line denotes nominal gravity magnitude

Figures 6-9 plot the estimates for the gyroscope bias, the magnetometer parameters and the gravity/magnetic vectors, respectively. All estimates converge within 30 seconds. The final gyroscope bias estimate computed by EKF in Test #1 is listed in Table II, as well as those for other three tests. It shows that the proposed EKF algorithm is able to estimate the gyroscope bias in dynamic condition up to 0.03 deg/s (taking the value by still averaging as reference), which is over three times better than what was achieved in [1]. In order to evaluate the quality of the obtained magnetometer parameters, we use the final estimate of $S$ and $h$ to calculate the magnitude of the hand-tumbled magnetometer measurement by $\|m^i\| = \|S^{-1}(y_m - h)\|$. Figure 10 gives the histogram for Test #1 of the magnitude discrepancy from 1 and the fitted normal distribution (mean $8\times 10^{-4}$, standard variance 0.005), compared with that of the raw magnetometer measurements. Additionally, we decompose the magnetometer matrix by $S^{-1} = C_m^b R$ into the cross-sensor and intrinsic parameters, as listed in Table II for all tests. Regarding the cross-sensor misalignment angle, the discrepancy between Tests #1-#2 or Tests #3-#4 is reduced to less than 0.2 degrees, in contrast to 0.4 degrees in [1]. In contrast to [1] that puts equal weights on sub-functions of the total cost function, Kalman filters are able to yield the state estimate naturally weighted by the propagated state covariance. It is the probable reason better results have been obtained herein.

Note that as being affected by the attached coin, the real magnetometer parameters deviate considerably from the given initial values in Tests #3-#4, particularly the magnetometer matrix $S$. Therefore we process the datasets #3-#4 by the EKF twice, in which the second EKF run uses as initial setting the estimated magnetometer matrix $S$ by the first EKF run while the other initial settings keep unchanged. All results of Tests #3-#4 in Tables I-II are the final estimates of the second EKF run. Figures 11-13 plot the estimates for the gyroscope bias and the magnetometer parameters in the second EKF run for Test #3. Table III compares the magnetometer ANIS metrics for all tests.

From the standpoint of the calibration task, the gravity/magnetic vector estimates in Fig. 8 are auxiliary states. They can be used to derive the local magnetic inclination as $90 - \cos^{-1}\left(m_{rs}^{iT} g^i / \|g^i\|\right)$ in degrees. The result is 43.01 degrees, in accordance with the value (43.14 degrees) by the World Magnetic Model at the test site. They actually serve as pillar reference vectors for the inertial attitude estimate. The Euler angle discrepancy between the inertial attitude estimate and the gyroscope-maintained relative attitude (computed by integrating the gyroscope measurements after subtracting the still-averaging bias) for Test #1 and Test #3 is given in Figure 14. The angle discrepancy is about 0.5 degrees in Test #1 and about 1 degree in Test #3. This result indicates a quite good consistency considering that the gyroscope-maintained relative attitude is only 2 degrees/5 minutes in accuracy (Fig. 4).

Theorem 2.1 tells that the problem could be solved using only magnetometer/gyroscope measurement information, so we check how the algorithm behaves if the accelerometer measurement is abandoned. The main results are summarized in Tables I-III and Fig. 15 plots the Euler angle discrepancy for Test #1 and Test #3. Although the magnetometer ANIS becomes significantly larger in the first EKF run of Test #3 (Table III), the proposed algorithm yields almost equivalent result in the gyroscope bias (Table I) and magnetometer parameters (Table II) and performs one time worse in attitude (Fig. 15 vs. Fig. 14). These observations show that the accelerometer information is of little help in magnetometer calibration and alignment to inertial sensors under motion conditions, and yet is quite helpful in mitigating attitude errors. As shown in Fig. 16, the accelerometer output taken valid by the EKF (indicated by red dots) is only about 5% of the total hand-tumbling data in Test #1 and Test #3. Enlarging the magnitude discrepancy $T_{md}$ by ten times brings the percentage of valid points up to 40-50%, but little estimate improvement has been obtained.

## IV. CONCLUSIONS

Modules or chips, consisting of triads of inertial and magnetic sensors, are enormously used in scientific or consumer devices. Convenient, reliable and accurate mutual calibrations of these sensors are a prerequisite for any practical use. Magnetometer calibration (and its alignment to inertial sensors) is usually achieved by heavily relying on accelerometer measurements at still, and thus are infeasible in motion conditions. This paper formulates the problem of magnetometer calibration and alignment to inertial sensors as a state estimation problem, collectively and recursively solving the magnetometer intrinsic calibration and cross-sensor calibration, as well as the gyroscope bias estimation. Sufficient conditions are derived for the problem to be globally observable, even when the accelerometer information is not used at all. An extended Kalman filter is designed to implement the state estimation and comprehensive test data results show the superior performance of the proposed approach. It is found that the accelerometer information, though helpful in mitigating attitude errors, is of little benefit in magnetometer calibration and alignment to inertial sensors under motion conditions.

The datasets used are available online at https://www.researchgate.net/profile/Yuanxin_Wu/contributions.

## REFERENCES

[1]  Y. Wu and S. Luo, "On Misalignment between Magnetometer and Inertial Sensors," *IEEE Sensors Journal,* vol. 16, pp. 6288-6297, 2016.
[2]  J. F. Vasconcelos*, et al.*, "Geometric Approach to Strapdown Magnetometer Calibration in Sensor Frame," *IEEE Trans. on Aerospace and Electronic Systems,* vol. 47, pp. 1293–1306, 2011.
[3]  J. C. Springmann and J. W. Cutler, "Attitude-Independent Magnetometer Calibration with Time-Varying Bias," *Journal of Guidance, Control, and Dynamics,* vol. 35, pp. 1080-1088, 2012.
[4]  J. Metge*, et al.*, "Calibration of an inertial-magnetic measurement unit without external equipment, in the presence of dynamic magnetic disturbances," *Measurement Science and Technology,* vol. 25, 2014.
[5]  R. Alonso and M. D. Shuster, "TWOSTEP: a fast robust algorithm for attitude-independent magnetometer-bias determination," *The Journal of the Astronautical Sciences,* vol. 50, pp. 433-451, 2002.
[6]  R. Alonso and M. D. Shuster, "Complete linear attitude-independent magnetometer," *The Journal of the Astronautical Sciences,* vol. 50, pp. 477-490, 2002.
[7]  J. Hol, "Sensor Fusion and Calibration of Inertial Sensors, Vision, Ultra-Wideband and GPS," Doctoral dissertation, Department of Electrical Engineering, Linköping University, 2011.
[8]  Y. Wu and W. Shi, "On Calibration of Three-axis Magnetometer," *IEEE Sensors Journal,* vol. 15, pp. 6424 - 6431, 2015.
[9]  B. Grandvallet*, et al.*, "Real-Time Attitude-Independent Three-Axis Magnetometer Calibration for Spinning Projectiles: A Sliding Window Approach," *IEEE Trans. on Control System and Technology,* vol. 22, pp. 255-264, 2014.
[10] J. L. Crassidis*, et al.*, "Real-Time Attitude-Independent Three-Axis Magnetometer Calibration," *Journal of Guidance, Control, and Dynamics,* vol. 28, pp. 115-120, 2005.
[11] X. Li and Z. Li, "A new calibration method for tri-axial field sensors in strap-down navigation systems," *Measurement Science & Technology,* vol. 23, p. 105105, 2012.
[12] Z. Zhang and G. Yang, "Micromagnetometer Calibration for Accurate Orientation Estimation," *IEEE Trans. on Biomedical Engineering,* vol. 62, pp. 553-560, 2015.
[13] M. Kok and T. B. Schon, "Maximum likelihood calibration of a magnetometer using inertial sensors," in *19th World Congress of the International Federation of Automatic Control (IFAC)*, Cape Town, South Africa, 2014.
[14] M. Kok and T. B. Schön, "Magnetometer Calibration Using Inertial Sensors," *IEEE Sensors Journal,* vol. 16, pp. 5679-5689, 2016.
[15] P. D. Groves, *Principles of GNSS, Inertial, and Multisensor Integrated Navigation Systems*, 2nd ed.: Artech House, Boston and London, 2013.
[16] D. H. Titterton and J. L. Weston, *Strapdown Inertial Navigation Technology*, 2nd ed.: the Institute of Electrical Engineers, London, United Kingdom, 2007.
[17] C.-T. Chen, *Linear System Theory and Design*, 3rd ed.: Rinehart and Winston, Inc., 1999.
[18] Y. Wu*, et al.*, "Observability of SINS Alignment: A Global Perspective," *IEEE Trans. on Aerospace and Electronic Systems,* vol. 48, pp. 78-102, 2012.
[19] A. Gelb, *Applied Optimal Estimation*. Cambridge, Mass.,: M.I.T. Press, 1974.
[20] J. K. Lee and E. J. Park, "A Fast Quaternion-Based Orientation Optimizer via Virtual Rotation for Human Motion Tracking," *IEEE Trans. on Biomedical Engineering,* vol. 56, pp. 1574-1582, 2009.
[21] A. M. Sabatini, "Quaternion-based extendedKalman filter for determining orientation by inertial and magnetic sensing," *IEEE Trans. on Biomedical Engineering,* vol. 53, pp. 1346–1356, 2006.
[22] S. J. Julier and J. J. LaViola, "On Kalman Filtering With Nonlinear Equality Constraints," *IEEE Trans. on Signal Processing,* vol. 55, pp. 2774-2784, 2007.
[23] R. Zanetti*, et al.*, "Norm-Constrained Kalman Filtering," *Journal of Guidance, Control, and Dynamics,* vol. 32, pp. 1458-1465, 2009.
[24] Y. Bar-Shalom*, et al.*, *Estimation with Applications To Tracking and Navigation*: John Wiley & Sons, 2001.

Table I. Gyroscope Bias Estimates (unit: deg/s)

|  | By Still Averaging | By EKF | By EKF (No Acc.) |
|---|---|---|---|
| **Test #1** | $[-0.195 \quad 0.168 \quad 0.256]^T$ | $[-0.221 \quad 0.171 \quad 0.250]^T$ | $[-0.216 \quad 0.165 \quad 0.254]^T$ |
| **Test #2** | $[-0.252 \quad 0.152 \quad 0.251]^T$ | $[-0.228 \quad 0.158 \quad 0.235]^T$ | $[-0.229 \quad 0.153 \quad 0.235]^T$ |
| **Test #3** | $[-0.228 \quad 0.147 \quad 0.246]^T$ | $[-0.247 \quad 0.144 \quad 0.245]^T$ | $[-0.253 \quad 0.144 \quad 0.250]^T$ |
| **Test #4** | $[-0.237 \quad 0.153 \quad 0.250]^T$ | $[-0.227 \quad 0.146 \quad 0.245]^T$ | $[-0.231 \quad 0.149 \quad 0.243]^T$ |

Table II. Magnetometer Parameter Estimate

| Test | Intrinsic Parameter ($\mathbf{R}, \mathbf{h}$) | | Cross-sensor Parameter $\mathbf{C}_b^m$ (in Euler angles, deg) | |
|---|---|---|---|---|
|  | EKF | EKF (No Acc.) | EKF | EKF (No Acc.) |
| **#1** | $\begin{bmatrix} 1.0021 & -0.0039 & 0.0011 \\ 0 & 0.9969 & 0.0019 \\ 0 & 0 & 1.0058 \end{bmatrix}$ $[-0.5018 \quad 0.0421 \quad 0.2379]^T$ | $\begin{bmatrix} 1.0023 & -0.0050 & 0.0010 \\ 0 & 0.9972 & 0.0016 \\ 0 & 0 & 1.0060 \end{bmatrix}$ $[-0.5017 \quad 0.0428 \quad 0.2378]^T$ | $\begin{bmatrix} 0.005 \\ 0.014 \\ -0.120 \end{bmatrix}$ | $\begin{bmatrix} 0.014 \\ 0.003 \\ -0.143 \end{bmatrix}$ |
| **#2** | $\begin{bmatrix} 1.0019 & -0.0059 & 0.0006 \\ 0 & 0.9979 & 0.0010 \\ 0 & 0 & 1.0047 \end{bmatrix}$ $[-0.5013 \quad 0.0415 \quad 0.2374]^T$ | $\begin{bmatrix} 1.0020 & -0.0059 & 0.0006 \\ 0 & 0.9979 & 0.0008 \\ 0 & 0 & 1.0047 \end{bmatrix}$ $[-0.5013 \quad 0.0416 \quad 0.2372]^T$ | $\begin{bmatrix} -0.062 \\ 0.005 \\ -0.118 \end{bmatrix}$ | $\begin{bmatrix} -0.071 \\ 0.005 \\ -0.117 \end{bmatrix}$ |
| **#3** | $\begin{bmatrix} 1.2668 & 0.3130 & -0.4191 \\ 0 & 1.5174 & 0.3419 \\ 0 & 0 & 0.8240 \end{bmatrix}$ $[-0.6288 \quad -0.1026 \quad -0.2513]^T$ | $\begin{bmatrix} 1.2656 & 0.3134 & -0.4182 \\ 0 & 1.5169 & 0.3416 \\ 0 & 0 & 0.8238 \end{bmatrix}$ $[-0.6289 \quad -0.1027 \quad -0.2508]^T$ | $\begin{bmatrix} 16.272 \\ 23.944 \\ 10.069 \end{bmatrix}$ | $\begin{bmatrix} 16.265 \\ 23.873 \\ 10.071 \end{bmatrix}$ |
| **#4** | $\begin{bmatrix} 1.2649 & 0.3150 & -0.4177 \\ 0 & 1.5204 & 0.3401 \\ 0 & 0 & 0.8251 \end{bmatrix}$ $[-0.6307 \quad -0.1049 \quad -0.2446]^T$ | $\begin{bmatrix} 1.2651 & 0.3147 & -0.4178 \\ 0 & 1.5202 & 0.3402 \\ 0 & 0 & 0.8252 \end{bmatrix}$ $[-0.6307 \quad -0.1049 \quad -0.2447]^T$ | $\begin{bmatrix} 16.253 \\ 23.762 \\ 10.056 \end{bmatrix}$ | $\begin{bmatrix} 16.262 \\ 23.770 \\ 10.047 \end{bmatrix}$ |

Table III. Magnetometer ANIS Metrics for All Tests

| Test | Mag. ANIS | Mag. ANIS (No Acc.) |
|---|---|---|
| #1 | 3.56 | 3.61 |
| #2 | 3.74 | 3.77 |
| #3 ($1^{st}$ run, $2^{nd}$ run) | 4.14, 4.35 | 42.38, 4.16 |
| #4 ($1^{st}$ run, $2^{nd}$ run) | 6.50, 4.99 | 5.74, 5.02 |